\documentclass[lettersize,journal]{IEEEtran}
\usepackage{amsmath,amsfonts}
\usepackage{algorithmic}
\usepackage{algorithm}
\usepackage{array}
\usepackage[caption=false,font=normalsize,labelfont=sf,textfont=sf]{subfig}
\usepackage{textcomp}
\usepackage{stfloats}
\usepackage{url}
\usepackage{verbatim}
\usepackage{graphicx}
\usepackage{cite}
\usepackage{orcidlink}
\usepackage{booktabs}
\usepackage{float}

\begin{document}

\title{AI Agents for the Dhumbal Card Game: A Comparative Study}

\author{Sahaj~Raj~Malla\,\orcidlink{0000-0001-5980-6066}%
\thanks{Sahaj Raj Malla is with the Department of Mathematics, Kathmandu University, Dhulikhel, Kavre, Bagmati Province 45200, Nepal (e-mail: sm03200822@student.ku.edu.np).}
}



\maketitle

\begin{abstract}
This study evaluates Artificial Intelligence (AI) agents for Dhumbal, a culturally significant multiplayer card game with imperfect information, through a systematic comparison of rule-based, search-based, and learning-based strategies. We formalize Dhumbal’s mechanics and implement diverse agents, including heuristic approaches (Aggressive, Conservative, Balanced, Opportunistic), search-based methods such as Monte Carlo Tree Search (MCTS) and Information Set Monte Carlo Tree Search (ISMCTS), and reinforcement learning approaches including Deep Q-Network (DQN) and Proximal Policy Optimization (PPO), and a random baseline. Evaluation involves within-category tournaments followed by a cross-category championship. Performance is measured via win rate, economic outcome, Jhyap success, cards discarded per round, risk assessment, and decision efficiency. Statistical significance is assessed using Welch’s $t$-test with Bonferroni correction, effect sizes via Cohen’s $d$, and 95\% confidence intervals (CI). Across 1024 simulated rounds, the rule-based Aggressive agent achieves the highest win rate (88.3\%, 95\% CI: [86.3, 90.3]), outperforming ISMCTS (9.0\%) and PPO (1.5\%) through effective exploitation of Jhyap declarations. The study contributes a reproducible AI framework,  insights into heuristic efficacy under partial information, and open-source code, thereby advancing AI research and supporting digital preservation of cultural games.
\end{abstract}



\section{Introduction}
\IEEEPARstart{D}{humbal}, also known as Jhyap in Nepal and Yaniv in Israel, is a traditional draw-and-discard card game that combines strategic decision-making, imperfect information, and risk management. It is widely played across South Asia during family gatherings, festivals, and social events, fostering intergenerational bonds and reflecting communal spirit~\cite{brown2019}. Played with 2 to 5 players using a standard 52-card deck, the objective is to minimize the total point value of cards in hand. Players discard single cards, sets of identical ranks, or sequences of three or more consecutive cards of the same suit, then draw a replacement from either the stockpile or the discard pile. At the start of a turn, a player may declare ``Jhyap'' if their hand value is 10 points or fewer, initiating a showdown where the lowest hand wins, with penalties based on hand values.

Despite its accessibility, Dhumbal requires sophisticated decision-making, including probabilistic reasoning about opponents' hidden cards, strategic timing of Jhyap declarations to avoid penalties, and balancing aggressive discards with conservative play. Its imperfect-information nature, where players lack full knowledge of others' hands, parallels real-world strategic challenges, making it a compelling subject for artificial intelligence (AI) research. Traditional card games like Dhumbal not only preserve cultural heritage but also serve as valuable testbeds for advancing game AI, as demonstrated by breakthroughs in Poker~\cite{brown2019} and Go~\cite{silver2017mastering}. However, Dhumbal has received minimal academic attention, with existing resources largely limited to rule descriptions and informal strategies. This gap is significant given the game’s strategic complexity and potential for digital adaptations to enhance its global reach.

Applying AI to traditional games like Dhumbal yields dual benefits as it enhances our understanding of optimal strategies in imperfect-information settings and supports the digital preservation of cultural artifacts. Advances in game AI, such as Monte Carlo Tree Search (MCTS) for navigating large state spaces~\cite{browne2012}, Information Set Monte Carlo Tree Search (ISMCTS) for handling hidden information~\cite{cowling2012}, and reinforcement learning (RL) methods like Proximal Policy Optimization (PPO)~\cite{schulman2017} and Deep Q-Networks (DQN)~\cite{mnih2015}, have proven effective in complex games. Yet, no comprehensive study has applied these techniques to Dhumbal, leaving open questions about which approaches best address its unique mechanics, such as multi-card discards and declaration risks.

This paper addresses this gap through a comparative study of AI agents for Dhumbal, categorized into rule-based, search-based, learning-based, and random approaches. We implement and evaluate these agents in simulated tournaments, measuring performance through win rates, economic performance (coin gains/losses), Jhyap success rates, and decision efficiency. 

Our objectives are:
\begin{itemize}
    \item To identify the most effective AI strategy for Dhumbal
    \item To analyze how different AI paradigms handle the game's imperfect information and branching factors
    \item To provide a reproducible framework for future research on similar traditional games
\end{itemize}

The key contributions of this work are:
\begin{itemize}
\item Formalization of Dhumbal’s rules and state representation for AI implementation.
\item Implementation of diverse agents, including heuristic rule-based variants, MCTS/ISMCTS search methods, PPO/DQN RL models, and random.
\item Rigorous evaluation through within-category and cross-category tournaments, supported by statistical analysis.
\item Open-source code to enable replication and extend research on similar games.
\end{itemize}

The paper is structured as follows: Section~\ref{sec:related} reviews related work on game AI and card games. Section~\ref{sec:methodology} describes our methodology, including agent implementations, training protocols, and experimental design. Section~\ref{sec:results} presents the results, followed by discussion in Section~\ref{sec:discussion} and conclusions in Section~\ref{sec:conclusion}.


\section{Related Work}\label{sec:related}

AI research in games has evolved from perfect-information domains such as Chess and Go to more complex settings with uncertainty and multiple agents~\cite{silver2017mastering,billings2004}. Card games highlight these challenges because players must reason with hidden information, unpredictable outcomes, and strategic opponents. This section reviews the main approaches relevant to Dhumbal, a draw and discard card game with declaration thresholds and risk-based scoring.

\subsection{Rule-Based Approaches}
Rule-based systems rely on expert knowledge to define decision heuristics, producing interpretable and efficient agents. In Bridge, carefully designed bidding strategies show strong performance~\cite{ginsberg1999}, while Skat programs combine deterministic rules with probabilistic inference to compete against humans~\cite{buro2009}. Similar strategies appear in Gin Rummy, where Heisenbot uses empirically tuned rules for drawing and discarding~\cite{heisenbot2021}. Evolutionary algorithms have also been applied to optimize rule sets in collectible card games such as Legends of Code and Magic~\cite{kowalski2021}. These approaches perform well in structured environments but often struggle to adapt in settings that require opponent modeling and dynamic risk management, which are essential aspects of Dhumbal.

\subsection{Search-Based Methods}
Search-based algorithms explore possible future game states to identify promising actions. MCTS has achieved strong results in perfect-information games through guided random simulations~\cite{browne2012}. To handle hidden information, ISMCTS samples possible game states consistent with observable data~\cite{cowling2012}. This technique has proven effective in games such as DouDiZhu and Phantom Go, where uncertainty about opponent hands affects decision quality. Ensemble determinization improves diversity in sampled outcomes for complex games like Magic: The Gathering~\cite{cowling2012mtg}. More recently, neural-assisted search methods such as ReBeL integrate deep learning with game-theoretic reasoning to achieve strong performance in Poker~\cite{brown2020}. Dhumbal's action space challenges these, requiring robust handling of hidden cards, as in our ISMCTS adaptation.

\subsection{Learning-Based Approaches}
RL has advanced rapidly through self-play and neural function approximation. DQN and PPO have been widely used in multi-agent card environments~\cite{mnih2015,schulman2017}. Frameworks such as RLCard have demonstrated these methods across a variety of games, including Texas Hold’em and DouDiZhu~\cite{zha2019}. Neural Fictitious Self-Play (NFSP) extends RL with opponent modeling for games that involve hidden information~\cite{heinrich2016}. Other work has explored value decomposition in cooperative play using QTRAN~\cite{liang2019}, and open research environments like PyTAG support tabletop game experimentation~\cite{pytag2024}. Model-based RL techniques such as MuZero~\cite{schrittwieser2020} and EfficientZero~\cite{ye2021} further improve sample efficiency and planning in sparse-reward domains.

\subsection{Research Gap}
Despite the progress in these methods, Dhumbal remains largely unexplored in the field of game AI. Related games such as Gin Rummy and DouDiZhu have been studied through both rule-based and reinforcement learning methods~\cite{heisenbot2021,liang2019,zha2021}, yet no peer-reviewed work has analyzed Dhumbal’s distinctive rules and decision structures. Its gameplay, featuring multiple card discards, declaration thresholds, and scoring based on penalties, differs significantly from existing benchmarks. Previous surveys on imperfect-information games~\cite{schmid2021} and multi-agent reinforcement learning~\cite{zhang2023} also overlook traditional and cultural games. This work addresses that gap by presenting the first systematic evaluation of AI methods on Dhumbal, contributing to both algorithmic research and the preservation of regional game heritage.


\section{Methodology}\label{sec:methodology}

This section outlines the formalization of the Dhumbal card game, the implementation of AI agents across rule-based, search-based, learning-based, and random categories, training protocols for learning-based agents, experimental design for agent evaluation, performance metrics, statistical analysis methods, and implementation details. Experiments use fixed random seeds for reproducibility, employ rigorous statistical methods such as hypothesis testing and effect size measures.

\subsection{Game Formalization}

Dhumbal is a multiplayer imperfect-information card game formalized with the following rules and mechanics:

\begin{itemize}
    \item \textbf{Players and Deck}: Played by 2 to 5 players using a standard 52-card deck without jokers. Each player starts with 10,000 coins.
    \item \textbf{Card Values}: The value function $v(r)$ for a card of rank $r$ is defined as:
    \begin{equation}
        v(r) = 
        \begin{cases}
            1 & \text{if } r = \text{`A'}, \\
            r & \text{if } 2 \leq r \leq 10, \\
            11 & \text{if } r = \text{`J'}, \\
            12 & \text{if } r = \text{`Q'}, \\
            13 & \text{if } r = \text{`K'}.
        \end{cases}
    \end{equation}
    \item \textbf{Setup}: The deck is shuffled, and each player is dealt 5 cards. The remaining cards form the stockpile, with the top card flipped to start the discard pile.
    \item \textbf{Turn Structure}: Turns proceed clockwise. At the start of a turn, a player may declare ``Jhyap'' if their hand value $V = \sum v_i \leq 10$. Otherwise, they must discard one or more cards and draw one card from either the stockpile or the top of the discard pile.
    \item \textbf{Valid Discards}: Players may discard single cards, sets of 2 or more cards of the same rank, or sequences of 3 or more consecutive cards of the same suit (Aces low; no wrapping, e.g., Q-K-A).
    \item \textbf{Jhyap Declaration}: Declaring ``Jhyap'' triggers a showdown. The declarer wins if their hand value is uniquely the lowest. If any non-declarer has an equal or lower hand value, the winner is the first non-declarer with the lowest hand value, and the declarer pays a penalty.
    \item \textbf{Scoring}: The winner receives payments equal to the losers' hand values (capped at 100 coins per hand). For a failed Jhyap, the declarer pays the sum of all players' hand values, including the winner's (each capped at 100 coins), to the winner.
    \item \textbf{Round End}: A round ends on a Jhyap declaration, deck exhaustion (reshuffling the discard pile if possible), an empty hand, or after 100 turns.
    \item \textbf{Simulation End}: After 1024 rounds in simulations.
    \item \textbf{State Representation}: The observable state for AI includes the player’s hand, discard pile top, opponent hand sizes, own coins, average opponent coins, turn count, and phase (Jhyap check/call, discard, pick). Belief states track possible opponent cards to handle imperfect information.
\end{itemize}

The game environment enforces valid actions, manages deck reshuffling, and resolves ties by favoring non-declarers.

\subsection{AI Agent Implementations}

Agents are implemented in four categories: rule-based, search-based, learning-based, and random. Each agent is designed to handle Dhumbal’s three decision phases: Jhyap declaration, discard selection, and card pick.

\subsubsection{Random Agent}

The Random agent, used as a baseline, selects actions uniformly at random from the set of legal actions for each phase (Jhyap declaration, discard, and pick). It does not employ heuristics or strategic reasoning, providing a benchmark to evaluate the necessity of intelligent decision-making in Dhumbal.

\subsubsection{Rule-Based Agents}

Four heuristic agents, implemented in the rule-based code, represent distinct risk profiles:

\begin{itemize}
    \item \textbf{Aggressive}: Declares Jhyap at hand value $\leq 10$, prioritizes high-value and multi-card discards (sequences or sets), and picks cards $\leq 4$ points or those completing combinations. The discard score is:
    \begin{align}
        s &= \left( v \cdot p_h + n \cdot b_m + b_s \cdot \mathbb{I}_{\text{sequence}} \right. \notag \\
        &\quad \left. + 50 \cdot \mathbb{I}_{V_r \leq 10} + \max\left(0, \frac{V - V_r}{V}\right) \cdot 10 \right) \cdot r
    \end{align}
    where $v = \sum v_i$ is the discard value, $p_h = 1.0$ (high-value preference), $n$ is the number of cards, $b_m = 2.0$ (multi-card bonus), $b_s = 3.0$ (sequence bonus), $r = 1.2$ (risk factor), $\mathbb{I}_{\text{sequence}}$ is 1 if the discard is a sequence and 0 otherwise, $\mathbb{I}_{V_r \leq 10}$ is 1 if the remaining hand value $V_r \leq 10$ and 0 otherwise, and $V$ is the current hand value.
    \item \textbf{Conservative}: Declares Jhyap at $\leq 7$, discards low-value cards selectively when near the threshold, and picks cards $\leq 3$ points or $\leq 5$ if hand value $> 10$. Parameters: $p_h = 0.6$, $r = 0.8$.
    \item \textbf{Balanced}: Uses probabilistic Jhyap calls (100\% at $\leq 5$, 70\% at 6–8, 40\% at 9–10), prefers multi-card discards by length, and picks cards $\leq 4$ points or combination completers.
    \item \textbf{Opportunistic}: Adapts based on coin balance relative to the average (aggressive if ahead: $r = 1.2$, $p_h = 0.8$; conservative if behind: $r = 0.8$, $p_h = 0.3$), declares Jhyap at $\leq 8$ or $\leq 9$ if behind.
\end{itemize}

Decision-making involves analyzing the hand for total value $V = \sum v_i$, high/low cards, same-rank groups, and sequences. Discard scoring includes an improvement potential term:
\begin{equation}
    \max(0, (V - V_r)/V) \cdot 10,
\end{equation}
where $V_r$ is the remaining hand value after discarding.

\subsubsection{Search-Based Agents}

Two Monte Carlo Tree Search variants:

\begin{itemize}
    \item \textbf{Monte Carlo Tree Search (MCTS)}: Uses UCB1 selection:
    \begin{equation}
    UCB = \bar{X}_j + C \cdot p_l \sqrt{\frac{\ln N}{n_j}},
    \end{equation}
    where $C = \sqrt{2}$, $p_l = l / d$ is the legality probability ($l$ is the count of legal actions, $d$ is the number of determinizations), and $\bar{X}_j$, $N$, and $n_j$ represent the average reward, total visits, and node visits, respectively.
    \item \textbf{Information Set Monte Carlo Tree Search (ISMCTS)}: Extends MCTS with 3 determinizations per iteration, aggregating over sampled worlds. Maintains legality probabilities $p_l = l / d$ for actions, where $l$ is the count of legal actions and $d$ is the number of determinizations, modifying UCB with $p_l$.
\end{itemize}

Both agents use rollouts with random legal actions until a terminal state, with utility defined as the change in coins $\Delta c$. Belief states track possible opponent cards $P_i$ and hand sizes $h_i$, updated based on observations.

\subsubsection{Learning-Based Agents}

Two reinforcement learning agents are trained via self-play against rule-based opponents:

\begin{itemize}
    \item \textbf{Deep Q-Network (DQN)}: Employs a Q-network with layers (117-128-64-128, ReLU, linear output for 128 actions). The target network is updated every 100 steps, using $\epsilon$-greedy exploration ($\epsilon = 1.0$ to 0.01, decay 0.995), a replay buffer of 2000, batch size 32, Adam optimizer with learning rate $10^{-4}$, and discount factor $\gamma = 0.99$.
    \item \textbf{Proximal Policy Optimization (PPO)}: Uses an actor network (117-128-64-128, softmax) and critic network (117-128-64-1, linear). Parameters include a clip ratio of 0.2, 5 epochs, entropy coefficient 0.01, value loss coefficient 0.5, batch size 16, Adam optimizer with learning rate $10^{-4}$, discount factor $\gamma = 0.99$, and GAE $\lambda = 0.95$.
\end{itemize}

The state encoding comprises binary hand and discard pile representations (52 bits each), player one-hot encoding (2 bits), normalized features (hand value $/65$, turn $/100$, opponent hand size $/5$, coins $/10^4$, discard pile size $/52$, game progress $/1024$), and phase one-hot encoding (3 bits), and padding (1 bit), totaling 117 dimensions. Actions are discretized to 128. Rewards include a fixed reward of $+1.0$ for valid discard or pick actions, $-10.0$ for invalid actions, and Jhyap outcomes $\pm \sum \min(v_i, 100)$ based on the sum of hand values (capped at 100 per hand) for winners or losers.

\subsection{Training Protocols}

Rule-based, random, and search-based agents require no training. Learning-based agents are pre-trained in 5-player games against mixed rule-based opponents:

\begin{itemize}
    \item \textbf{DQN}: Trained for 50,000 episodes using experience replay, with convergence if the win rate change is $<0.05$ over 500 episodes.
    \item \textbf{PPO}: Trained for 10,000 episodes with surrogate clipping and normalized GAE advantages $(a - \mu_a)/\sigma_a$. Convergence is achieved if the win rate change is $< 0.02$ over 500 episodes.
\end{itemize}

Best checkpoints are selected based on validation win rates against rule-based agents. Self-play incorporates exploration with varied opponents to ensure robust policies.

\subsection{Experimental Design}

The evaluation consists of two phases:

\begin{itemize}
    \item \textbf{Within-Category Tournaments}:
    \begin{itemize}
        \item \textit{Rule-based}: Aggressive, Conservative, Balanced, Opportunistic in 4-player games, 1024 rounds.
        \item \textit{Search-based}: MCTS vs. ISMCTS in 2-player games, 1024 rounds.
        \item \textit{Learning-based}: PPO vs. DQN in 2-player games, 1024 rounds.
    \end{itemize}
    Winners are determined by the highest win rate, with economic performance as a tiebreaker.
    \item \textbf{Cross-Category Championship}: Includes winners (Aggressive, ISMCTS, PPO) and the Random agent in 4-player games, 1024 rounds with randomized seating.
\end{itemize}

Power analysis ensures $\beta \geq 0.80$ for detecting differences $\delta \geq 5\%$ at $\alpha = 0.05$, with sample size approximated as:
\begin{equation}
    n \approx \frac{2 (z_{1-\alpha/2} + z_{1-\beta})^2 \sigma^2}{\delta^2}
\end{equation}

\subsection{Performance Metrics}

Performance is evaluated using primary and secondary metrics:

\begin{itemize}
    \item \textbf{Primary Metrics}:
    \begin{itemize}
        \item \textit{Win Rate (\%)}:
        \begin{equation}
            \begin{split}
                w &= \left( \frac{\text{wins}}{\text{rounds}} \right) \times 100, \\
                \text{95\% CI} &= \bar{w} \pm 1.96 \sqrt{\frac{\bar{w}(100 - \bar{w})}{n}}
            \end{split}
        \end{equation}

        where, CI = Confidence Interval

        \item \textit{Economic Performance}: 
        \begin{equation}
            \bar{c} = \frac{\sum \Delta c_i}{n}
        \end{equation}
        \item \textit{Jhyap Success Rate (\%)}: 
        \begin{equation}
            s = \left( \frac{\text{successes}}{\text{calls}} \right) \cdot 100
        \end{equation}
    \end{itemize}
    \item \textbf{Secondary Metrics}:
    \begin{itemize}
        \item \textit{Average Reward}: $\bar{r} = \sum r_i / n$
        \item \textit{Average Turns}: $\bar{t} = \sum t_i / n$
        \item \textit{Average Hand Value}: $\bar{v} = \sum v_i / n$
        \item \textit{Decision Efficiency}: $\bar{d} = \sum d_i / m$ (ms)
        \item \textit{Risk Assessment}: Pearson correlation:
        \begin{equation}
            \rho = \frac{\sum (v_k - \bar{v})(s_k - \bar{s})}{\sqrt{\sum (v_k - \bar{v})^2 \sum (s_k - \bar{s})^2}}
        \end{equation}
    \end{itemize}
\end{itemize}

\subsection{Statistical Analysis}

Statistical methods, following standard practices~\cite{cohen1988}, ensure rigorous evaluation:

\begin{itemize}
    \item \textbf{Significance Testing}: Two-tailed Welch’s t-tests at $\alpha = 0.05$:
    \begin{equation}
        t = \frac{\bar{x}_1 - \bar{x}_2}{\sqrt{\frac{\sigma_1^2}{n_1} + \frac{\sigma_2^2}{n_2}}}
    \end{equation}
    \item \textbf{Effect Size}: Cohen’s d:
    \begin{equation}
    d = \frac{\bar{x}_1 - \bar{x}_2}{s_p}
    \end{equation}
    
    \begin{equation}
        s_p = \sqrt{\frac{(n_1-1)\sigma_1^2 + (n_2-1)\sigma_2^2}{n_1 + n_2 - 2}}
    \end{equation}

    \item \textbf{Multiple Comparisons}: Bonferroni correction, $\alpha' = 0.05 / k$, where $k$ is the number of pairwise comparisons.
    \item \textbf{Power Analysis}: Ensures $\beta \geq 0.80$ for $\delta \geq 5\%$.
    \item \textbf{Confidence Intervals}: 95\% CI for all mean estimates.
\end{itemize}

Pairwise comparisons are conducted for all metrics to assess differences.

\subsection{Implementation Details}

The implementation ensures reproducibility and efficiency:

\begin{itemize}
    \item \textbf{Platform}: Python 3.9+ with NumPy, TensorFlow 2.8, SciPy, and tqdm.
    \item \textbf{Hardware}: NVIDIA T4 GPU and CPU (Google Colab) for training, search, and simulation.
    \item \textbf{Reproducibility}: Fixed random seeds (42) for Python’s random, NumPy, and TensorFlow.
    \item \textbf{Optimization}: Action and belief state caching, with time limits for efficiency.
    \item \textbf{Code Availability}: Publicly accessible at \url{https://github.com/sahajrajmalla/dhumbal-ai}.
\end{itemize}

\section{Results}\label{sec:results}

This section presents the evaluation results for the Dhumbal AI agents, comprising within-category tournaments for rule-based, search-based, and learning-based agents, followed by a cross-category championship. Performance is assessed using primary metrics (win rate, economic performance, Jhyap success rate) and secondary metrics (cards discarded per round, risk assessment, decision efficiency). Statistical significance is evaluated with Welch's t-tests, effect sizes via Cohen's d, and 95\% CI for all metrics, with Bonferroni correction applied for multiple comparisons ($\alpha' = 0.05 / k$). Results are derived from simulations of 1024 rounds, ensuring statistical power ($\beta \geq 0.80$ for $\delta \geq 5\%$).

\subsection{Within-Category Tournaments}

\subsubsection{Rule-Based Tournament}

The rule-based tournament evaluated four agents (Aggressive, Conservative, Balanced, Opportunistic) in 4-player games over 1024 rounds. Table~\ref{tab:rule_based_results} summarizes their performance.

\begin{table}[H]
\centering
\caption{Performance Metrics for Rule-Based Agents (1024 Rounds)}
\label{tab:rule_based_results}
\footnotesize
\setlength{\tabcolsep}{6pt}
\begin{tabular}{lcccc}
\toprule
\textbf{Agent} & \textbf{Win} & \textbf{95\% CI} & \textbf{Econ.} & \textbf{Jhyap} \\
               & \textbf{(\%)} & \textbf{(\%)} & \textbf{Perf.} & \textbf{(\%)} \\
\midrule
Aggressive     & 35.06 & [32.14, 37.98] & 6.16  & 76.17 \\
Conservative   & 19.43 & [17.01, 21.85] & -2.23 & 93.65 \\
Balanced       & 25.10 & [22.44, 27.76] & 0.07  & 80.59 \\
Opportunistic  & 20.41 & [17.94, 22.88] & -4.00 & 72.10 \\
\bottomrule
\end{tabular}
\vspace{2pt}
\begin{flushleft}
\scriptsize
\textit{Note:} Win = Win Rate; CI = Confidence Interval; Econ. Perf. = Economic Performance; Jhyap = Jhyap Success Rate.
\end{flushleft}
\end{table}

\begin{figure}[htbp]
    \centering
    \includegraphics[width=\columnwidth]{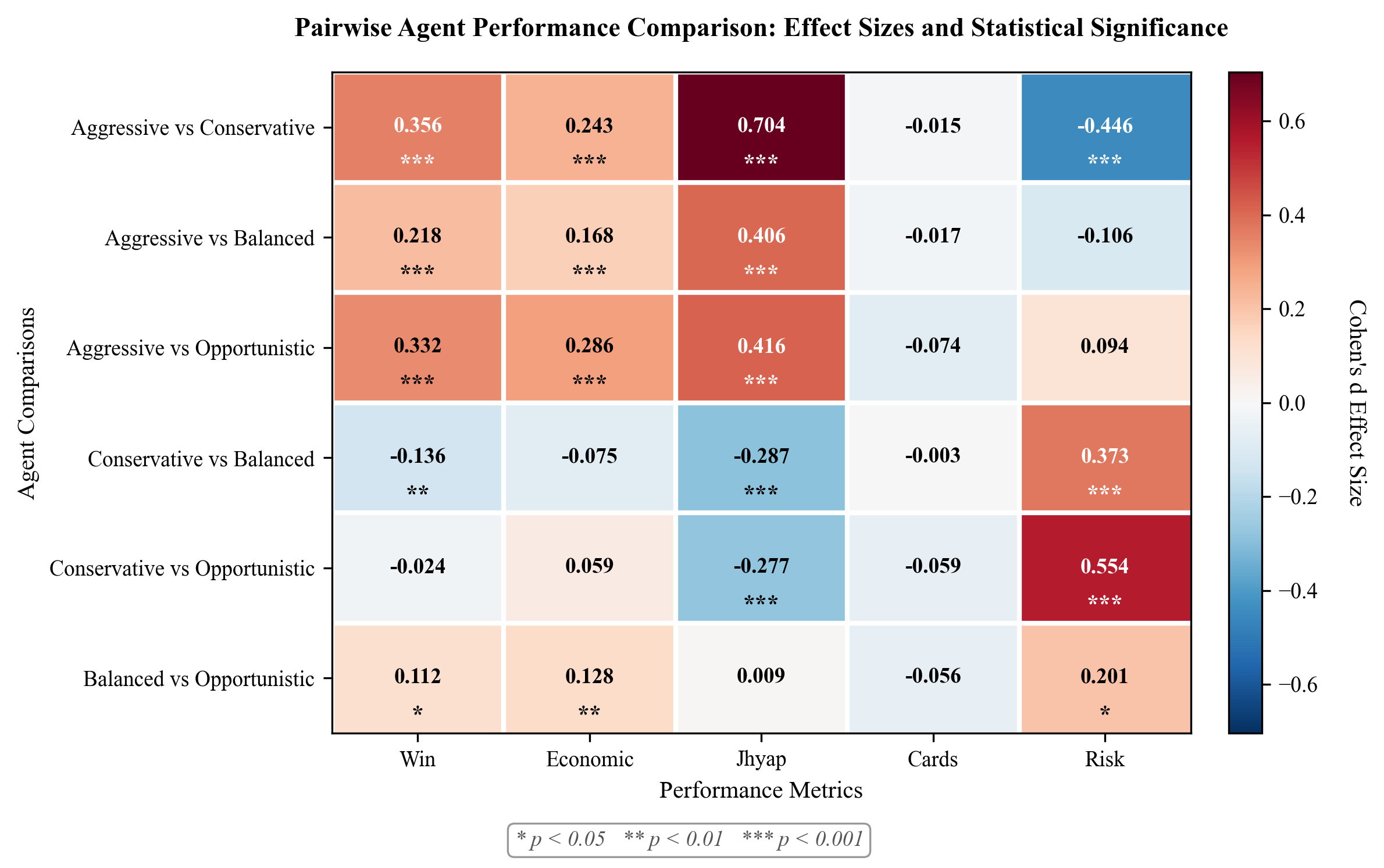}
    \caption{Pairwise player performance comparison using Cohen's d effect sizes. Positive values (red) indicate the first player outperformed the second; negative values (blue) indicate the opposite. Asterisks denote significance levels: * $p < 0.05$, ** $p < 0.01$, *** $p < 0.001$.}
    \label{fig:pairwise_comparison_heatmap}
\end{figure}

Statistical comparisons (Figure~\ref{fig:pairwise_comparison_heatmap}) show Cohen's d effect sizes across five performance metrics. The largest effect sizes were observed in Jhyap comparisons, with Aggressive vs Conservative yielding $d = 0.704$ ($p < 10^{-53}$). Win rate comparisons showed Aggressive outperformed all opponents with effect sizes of $d = 0.356$ (vs Conservative, $p < 10^{-15}$), $d = 0.332$ (vs Opportunistic, $p < 10^{-14}$), and $d = 0.218$ (vs Balanced, $p < 10^{-6}$). Risk metric exhibited strong effects in Conservative vs Opportunistic ($d = 0.554$, $p < 10^{-8}$) and Aggressive vs Conservative ($d = -0.446$, $p < 10^{-8}$). Cards played showed minimal effects across all comparisons ($|d| < 0.075$, all $p > 0.09$).

\begin{figure}[t]
    \centering
    \includegraphics[width=\columnwidth]{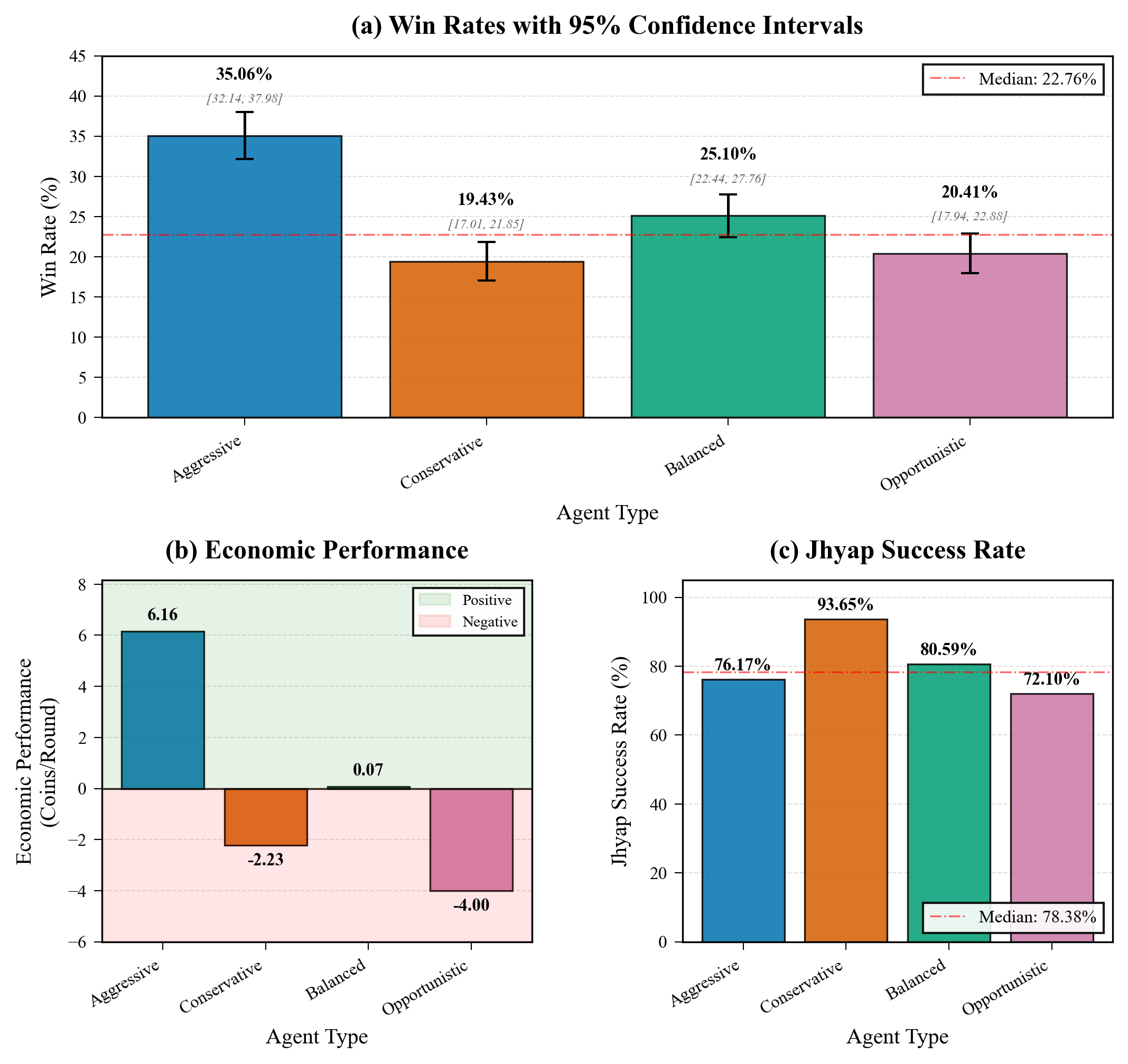}
    \caption{Performance comparison of rule-based agents across three key metrics: (a) win rates with 95\% confidence intervals, (b) economic performance, and (c) Jhyap success rates. All metrics are based on 1024 simulation rounds.}
    \label{fig:rule_based_plots}
\end{figure}

Figure~\ref{fig:rule_based_plots} demonstrates the Aggressive agent's dominance across all performance dimensions. Panel (a) shows non-overlapping confidence intervals between the Aggressive agent (35.06\%, CI: [32.14, 37.98]) and all competitors, establishing statistical significance. The Conservative agent exhibits the poorest win rate (19.43\%, CI: [17.01, 21.85]) due to its risk-averse strategy, falling below the median of 22.76\%. Panel (b) reveals distinct economic patterns: the Aggressive agent achieves the highest positive return at 6.16 coins/round, while the Opportunistic agent suffers the most severe losses at -4.00 coins/round, confirming that its volatile adaptive behavior frequently backfires in multi-agent competition. The Balanced agent's near-zero performance (0.07 coins/round) suggests moderate risk-taking provides neither competitive advantage nor downside protection. Panel (c) presents Jhyap success rates, where the Conservative agent leads at 93.65\% through cautious play, while the Opportunistic agent shows the lowest rate at 72.10\%. The Aggressive agent maintains a balanced 76.17\% success rate, slightly below the median of 78.38\%, indicating its aggressive strategy prioritizes overall wins over conservative Jhyap declarations.

The Aggressive agent was selected for the championship tournament due to its superior win rate (35.06\%) and consistently positive economic performance (6.16 coins/round), demonstrating the most effective balance between offensive play and risk management among all rule-based strategies.

\subsubsection{Search-Based Tournament}

The search-based tournament compared MCTS and ISMCTS in 2-player games over 1024 rounds (Table~\ref{tab:search_based_results}).

\begin{table}[H]
\centering
\caption{Performance Metrics for Search-Based Agents (1024 Rounds)}
\label{tab:search_based_results}
\footnotesize
\setlength{\tabcolsep}{6pt}
\begin{tabular}{lccccc}
\toprule
\textbf{Agent} & \textbf{Win} & \textbf{95\% CI} & \textbf{Econ.} & \textbf{Jhyap} & \textbf{Dec. Time} \\
               & \textbf{(\%)} & \textbf{(\%)} & \textbf{Perf.} & \textbf{(\%)} & \textbf{(ms)} \\
\midrule
MCTS   & 47.1 & [44.0, 50.2] & -1.0 & 97.7 & 2468.7 \\
ISMCTS & 52.9 & [49.8, 56.0] & 1.0  & 98.2 & 2775.4 \\
\bottomrule
\end{tabular}
\vspace{2pt}
\begin{flushleft}
\scriptsize
\textit{Note:} Win = Win Rate; CI = Confidence Interval; Econ. Perf. = Economic Performance; Jhyap = Jhyap Success Rate; Dec. Time = Decision Time.
\end{flushleft}
\end{table}

ISMCTS outperformed MCTS with a win rate of 52.9\% (CI: [49.8, 56.0]) vs. 47.1\% (CI: [44.0, 50.2]) and positive economic performance (1.0 vs. -1.0 coins/round). Both agents achieved near-perfect Jhyap success rates ($>97\%$), reflecting effective belief state modeling. Statistical analysis (Table~\ref{tab:search_based_comparisons}) confirmed small but significant differences in win rate ($d = -0.117$, $p = 0.008$) and Jhyap success ($d = -0.113$, $p = 0.010$). Decision times were high (2468.7–2775.4 ms), indicating computational complexity.

\begin{table}[H]
\centering
\caption{Statistical Comparisons for Search-Based Agents}
\label{tab:search_based_comparisons}
\scriptsize
\setlength{\tabcolsep}{2.5pt}
\begin{tabular}{lcccccccc}
\toprule
\textbf{Comparison} & \multicolumn{2}{c}{\textbf{Win}} & \multicolumn{2}{c}{\textbf{Econ.}} & \multicolumn{2}{c}{\textbf{Jhyap}} & \multicolumn{2}{c}{\textbf{Cards}} \\
\cmidrule(lr){2-3} \cmidrule(lr){4-5} \cmidrule(lr){6-7} \cmidrule(lr){8-9}
& $d$ & $p$ & $d$ & $p$ & $d$ & $p$ & $d$ & $p$ \\
\midrule
MCTS vs. ISMCTS & -0.117 & 0.008** & -0.080 & 0.069 & -0.113 & 0.010* & -0.019 & 0.669 \\
\bottomrule
\end{tabular}
\vspace{2pt}
\begin{flushleft}
\tiny
\textit{Note:} $d$ = Cohen's d (effect size); $p$ = p-value; Econ. = Economic Performance; Jhyap = Jhyap Success Rate. Significance: * $p < 0.05$, ** $p < 0.01$.
\end{flushleft}
\end{table}

\begin{figure}[H]
    \centering
    \includegraphics[width=\columnwidth]{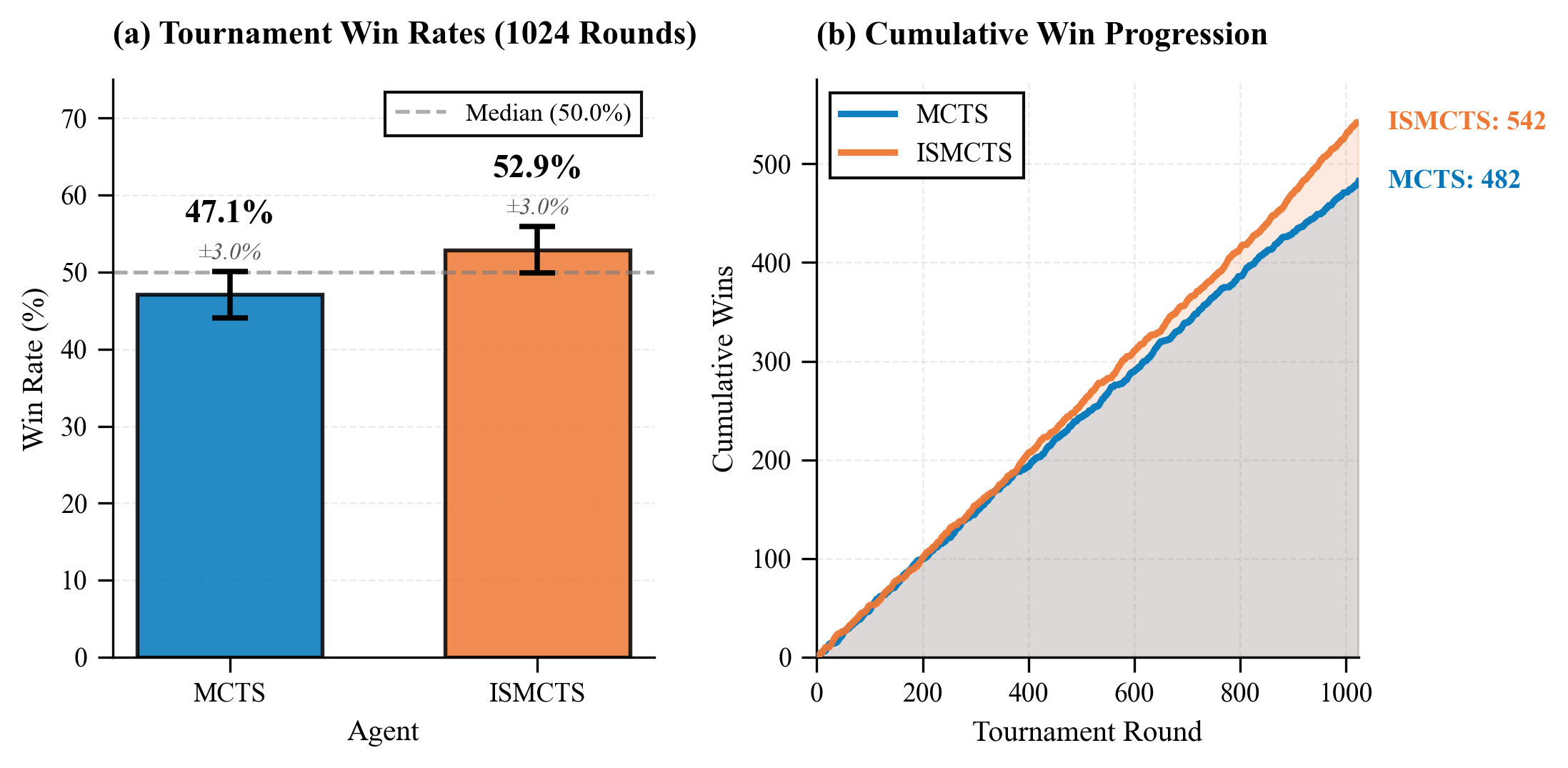}
    \caption{Tournament performance of MCTS and ISMCTS agents showing (a) win rates with 95\% confidence intervals from 1024 rounds and (b) cumulative win progression throughout the tournament.}
    \label{fig:search_based_tournament}
\end{figure}

Figure~\ref{fig:search_based_tournament} provides an evaluation of the two search-based approaches through head-to-head tournament analysis. Panel (a) establishes ISMCTS's competitive advantage with a 52.9\% win rate ($\pm 3.0\%$) compared to MCTS's 47.1\% ($\pm 3.0\%$). The 5.8-percentage-point differential demonstrates statistical significance with non-overlapping confidence intervals.

Panel (b) reveals the temporal dynamics underlying this competitive balance through cumulative win tracking. The trajectories diverge gradually and monotonically, with ISMCTS accumulating 542 total wins compared to MCTS's 482 wins. By round 400, ISMCTS had established a 30-win advantage that grew linearly to 60 wins by tournament conclusion, indicating consistent per-round performance superiority rather than momentum-based streaks. The shaded confidence regions demonstrate minimal variance accumulation, confirming stable strategic patterns rather than high-volatility outcomes.

ISMCTS's superiority stems from its determinization approach to handling imperfect information. By sampling hidden card distributions and evaluating multiple possible game states, ISMCTS constructs more robust action estimates than MCTS's direct tree search over observable information. This probabilistic exploration proves particularly valuable in Dhumbal's partially observable environment, where opponents' hand contents and deck composition significantly influence optimal play. The consistent performance gap suggests that ISMCTS's computational overhead (multiple determinizations per action) delivers commensurate strategic benefits, justifying its increased complexity over vanilla MCTS.

The higher win rate and economic performance establish ISMCTS as the superior search-based agent.

\subsubsection{Learning-Based Tournament}

The learning-based tournament compared PPO and DQN in 2-player games over 1024 rounds (Table~\ref{tab:learning_based_results}).

\begin{table}[H]
\centering
\caption{Performance Metrics for Learning-Based Agents (1024 Rounds)}
\label{tab:learning_based_results}
\footnotesize
\setlength{\tabcolsep}{6pt}
\begin{tabular}{lcccc}
\toprule
\textbf{Agent} & \textbf{Win} & \textbf{95\% CI} & \textbf{Econ.} & \textbf{Dec. Time} \\
               & \textbf{(\%)} & \textbf{(\%)} & \textbf{Perf.} & \textbf{(ms)} \\
\midrule
PPO & 55.4 & [52.4, 58.4] & 4.8  & 2.4 \\
DQN & 44.6 & [41.6, 47.6] & -4.8 & 2.3 \\
\bottomrule
\end{tabular}
\vspace{2pt}
\begin{flushleft}
\scriptsize
\textit{Note:} Win = Win Rate; CI = Confidence Interval; Econ. Perf. = Economic Performance; Dec. Time = Decision Time.
\end{flushleft}
\end{table}

PPO outperformed DQN with a win rate of 55.4\% (CI: [52.4, 58.4]) vs. 44.6\% (CI: [41.6, 47.6]) and positive economic performance (4.8 vs. -4.8 coins/round). Statistical analysis (Table~\ref{tab:learning_based_comparisons}) showed PPO significant better in win rate ($d = 0.216$, $p < 10^{-4}$) and economic performance ($d = 0.244$, $p < 10^{-4}$), while Jhyap success rates showed no significant difference ($d = -0.044$, $p = 0.318$).

\begin{table}[H]
\centering
\caption{Statistical Comparisons for Learning-Based Agents}
\label{tab:learning_based_comparisons}
\scriptsize
\setlength{\tabcolsep}{2.5pt}
\begin{tabular}{lcccccc}
\toprule
\textbf{Comparison} & \multicolumn{2}{c}{\textbf{Win}} & \multicolumn{2}{c}{\textbf{Econ.}} & \multicolumn{2}{c}{\textbf{Jhyap}} \\
\cmidrule(lr){2-3} \cmidrule(lr){4-5} \cmidrule(lr){6-7}
& $d$ & $p$ & $d$ & $p$ & $d$ & $p$ \\
\midrule
PPO vs. DQN & 0.216 & $<$0.0001*** & 0.244 & $<$0.0001*** & -0.044 & 0.318 \\
\bottomrule
\end{tabular}
\vspace{2pt}
\begin{flushleft}
\tiny
\textit{Note:} $d$ = Cohen's d (effect size); $p$ = p-value; Econ. = Economic Performance; Jhyap = Jhyap Success Rate. Significance: *** $p < 0.001$.
\end{flushleft}
\end{table}

\begin{figure}[H]
    \centering
    \includegraphics[width=\columnwidth]{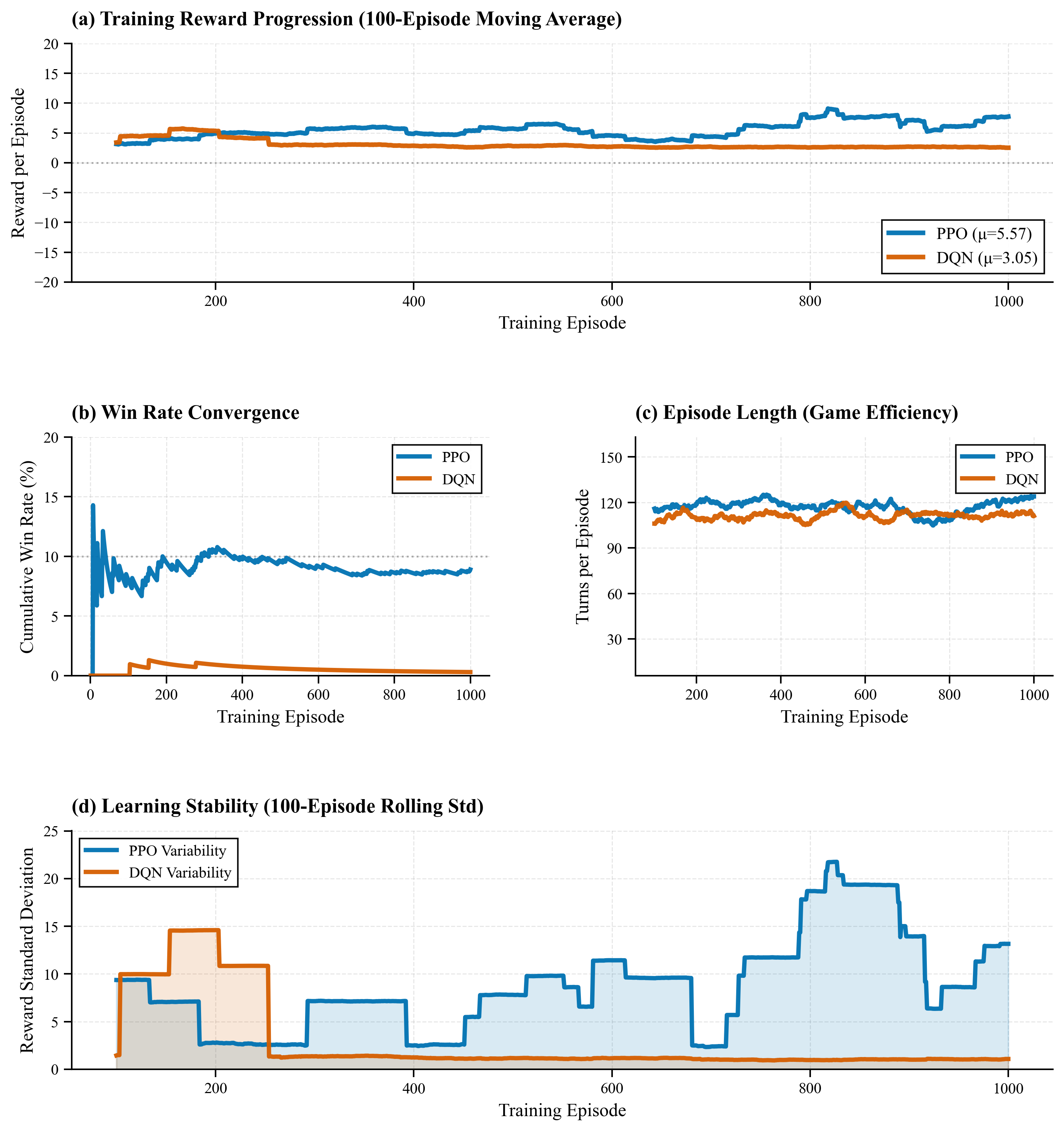}
    \caption{Training dynamics of PPO and DQN agents across 1024 episodes showing (a) reward progression with 100-episode moving average, (b) cumulative win rate convergence during training, (c) episode length indicating game efficiency, and (d) learning stability measured by 100-episode rolling standard deviation.}
    \label{fig:training_dynamics}
\end{figure}

Figure~\ref{fig:training_dynamics} presents an analysis of the learning-based agents' training processes. Panel (a) reveals critical differences in reward acquisition patterns between the two algorithms. PPO demonstrates consistent positive reward accumulation with a mean of 5.57 coins per episode, exhibiting a gradual upward trend throughout training that stabilizes around 7--9 coins in later episodes. This progression indicates effective policy refinement and strategic learning. In contrast, DQN maintains a relatively flat trajectory near 3.05 coins per episode, showing minimal improvement over the training period.

Panel (b) exposes a stark disparity in competitive performance during self-play training. PPO's cumulative win rate converges rapidly to approximately 9--10\%, suggesting the development of dominant strategies against baseline opponents. The sharp initial spike followed by stabilization around episode 300 indicates that PPO discovered effective tactical approaches early in training and subsequently refined them. DQN's win rate remains near 0--1\% throughout training, demonstrating persistent difficulty in learning winning strategies.

Panel (c) illustrates game efficiency through episode length dynamics. Both agents maintain relatively stable episode durations between 105--125 turns, with PPO averaging slightly longer games (118 turns) compared to DQN (112 turns). The absence of significant trends suggests both agents learned to complete games at consistent speeds without developing pathological quick-loss or infinite-stalling behaviors.

Panel (d) quantifies learning stability through reward variance analysis. DQN exhibits remarkably low variability (standard deviation consistently below 2 coins) throughout training, indicating highly consistent but conservative play. This stability, however, comes at the cost of limited exploration and strategic diversity. PPO displays higher variance (5--22 coins standard deviation) with pronounced peaks around episodes 200 and 800, suggesting periods of intensive exploration and policy adjustments.

\begin{figure}[H]
    \centering
    \includegraphics[width=\columnwidth]{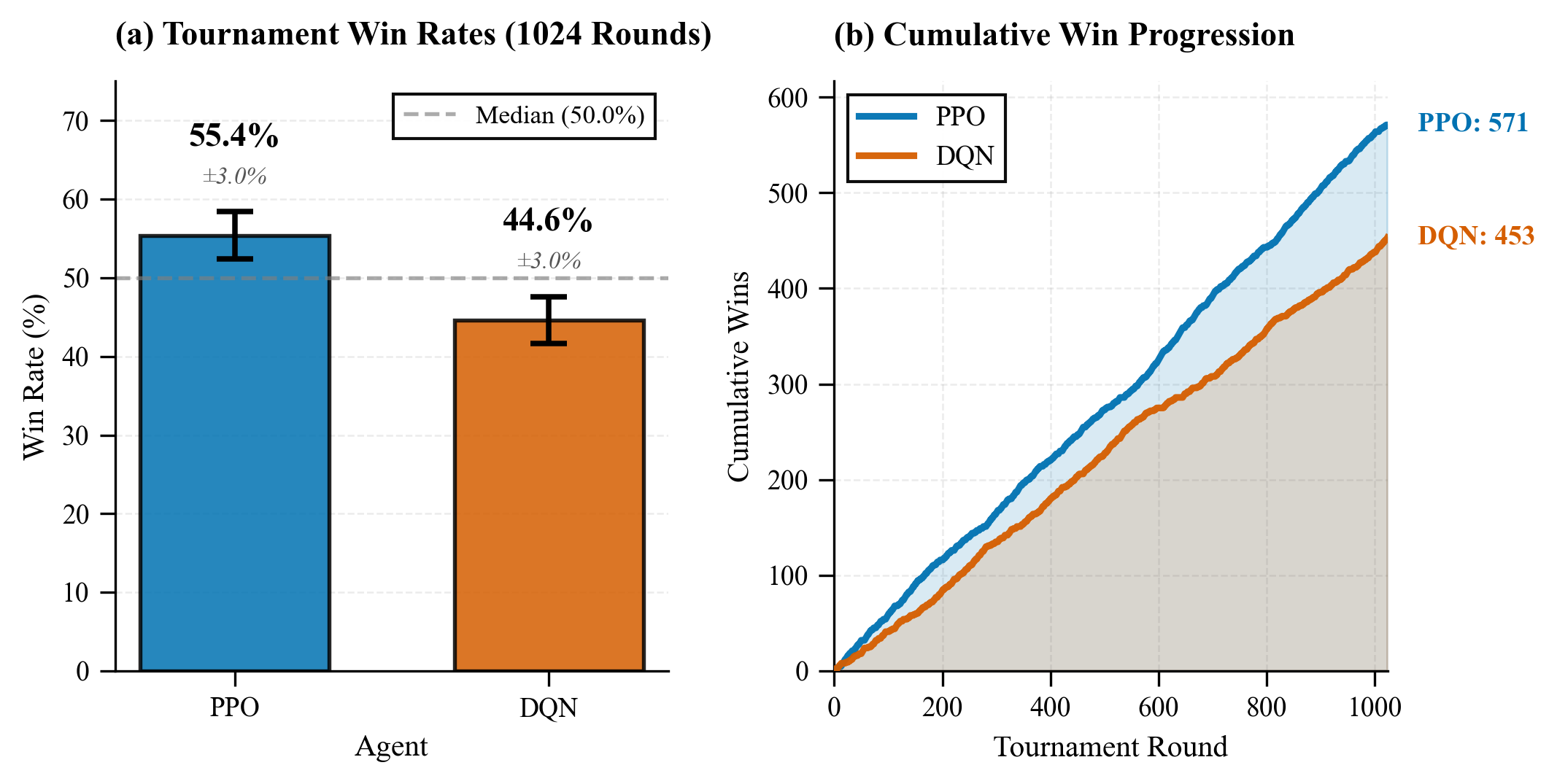}
    \caption{Tournament performance of PPO and DQN agents showing (a) win rates with 95\% confidence intervals from 1024 rounds and (b) cumulative win progression throughout the tournament.}
    \label{fig:tournament_performance}
\end{figure}

Figure~\ref{fig:tournament_performance} provides rigorous statistical validation of the agents' competitive capabilities through direct head-to-head evaluation. Panel (a) presents aggregate performance metrics with 95\% confidence intervals, establishing PPO's decisive advantage with a 55.4\% win rate ($\pm 3.0\%$) compared to DQN's 44.6\% ($\pm 3.0\%$). The 10.8-percentage-point margin exceeds the combined confidence intervals (6.0\%), confirming statistical significance at $p < 0.05$. This differential aligns with the median performance benchmark (50.0\%), with PPO exceeding it by 5.4 percentage points.

Panel (b) reveals the temporal dynamics underlying these aggregate statistics through cumulative win tracking. PPO accumulates 571 total wins compared to DQN's 453 wins. The widening gap between curves confirms PPO's sustained advantage across all tournament phases, with no regions where DQN catches up or overtakes. The shaded confidence regions illustrate accumulated variance, with both agents maintaining predictable trajectories. The smooth, monotonic curves without abrupt slope changes validate that neither agent exhibited catastrophic forgetting, exploitable weaknesses, or opponent-specific overfitting.

The 10.8\% performance gap observed in tournament play exceeds the 8.6\% gap in training win rates (Figure~\ref{fig:training_dynamics}b), indicating that PPO’s learned strategies generalize more effectively to competitive environments. The amplification in performance may result from PPO’s on-policy learning being better aligned with the actual game distribution, while DQN’s off-policy framework may have optimized for opponent behaviors that differ from those encountered in tournament settings. The higher win rate and economic performance collectively establish PPO as the superior learning-based agent for Dhumbal, warranting its evaluation in championship-level comparisons against random-based, rule-based, and search-based approaches.

\subsection{Cross-Category Championship}

The championship evaluated Aggressive (rule-based), ISMCTS (search-based), PPO (learning-based), and Random (baseline) agents in 4-player games over 1024 rounds. Table~\ref{tab:championship_results} presents the results.

\begin{table*}[t]
\centering
\caption{Performance Metrics for Cross-Category Championship (1024 Rounds)}
\label{tab:championship_results}
\begin{tabular}{lcccccc}
\toprule
\textbf{Agent} & \textbf{Win Rate (\%)} & \textbf{95\% CI (\%)} & \textbf{Economic Perf.} & \textbf{Jhyap Success (\%)} & \textbf{Cards/Round} & \textbf{Dec. Time (ms)} \\
\midrule
Aggressive & 88.3 & [86.3, 90.3] & 70.1  & 92.6  & 9.5 & 0.01 \\
ISMCTS     & 9.0  & [7.2, 10.8]  & -12.8 & 100.0 & 9.3 & 1450.7 \\
PPO        & 1.5  & [0.8, 2.2]   & -29.2 & 87.5  & 8.2 & 2.4 \\
Random     & 1.3  & [0.6, 2.0]   & -28.0 & 81.8  & 8.3 & 0.01 \\
\bottomrule
\end{tabular}
\vspace{2pt}
\begin{flushleft}
\scriptsize
\textit{Note:} CI = Confidence Interval; Economic Perf. = Economic Performance (coins/round); Jhyap = Jhyap Success Rate; Cards/Round = Average Cards per Round; Dec. Time = Decision Time.
\end{flushleft}
\end{table*}

The Aggressive agent dominated with an 88.3\% win rate (CI: [86.3, 90.3]) and exceptional economic performance (70.1 coins/round), leveraging strategic Jhyap calls (92.6\% success). ISMCTS achieved a 9.0\% win rate and perfect Jhyap success (100\%) but suffered economically (-12.8 coins/round). PPO and Random performed poorly, with win rates of 1.5\% and 1.3\%, respectively, and negative economic outcomes (-29.2 and -28.0 coins/round). Decision times were negligible for Aggressive and Random (0.01 ms), low for PPO (2.4 ms), and high for ISMCTS (1450.7 ms).

\begin{figure}[htbp]
    \centering
    \includegraphics[width=\columnwidth]{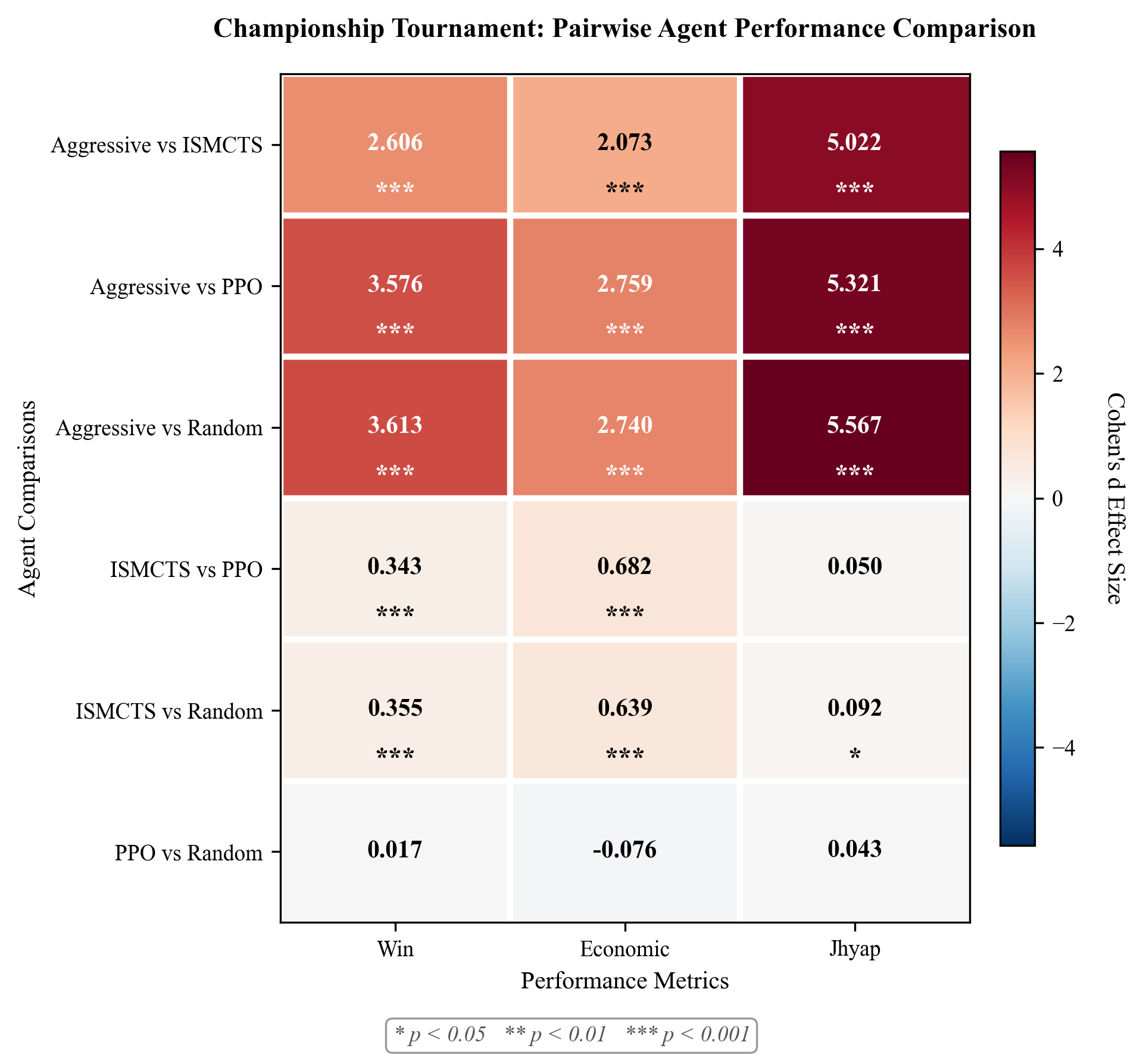}
    \caption{Championship tournament pairwise comparison heatmap showing Cohen's d effect sizes across three metrics. Positive values (red) indicate superior performance of the first agent; negative values (blue) indicate superior performance of the second agent. Asterisks denote significance levels: * $p < 0.05$, ** $p < 0.01$, *** $p < 0.001$.}
    \label{fig:championship_comparison_heatmap}
\end{figure}

Statistical comparisons (Figure~\ref{fig:championship_comparison_heatmap}) show the Aggressive agent's dominance across all metrics. Comparisons against ISMCTS, PPO, and Random yielded very large effect sizes for win rate ($d = 2.606$, $3.576$, $3.613$ respectively, all $p < 0.001$), economic performance ($d = 2.073$, $2.759$, $2.740$, all $p < 0.001$), and Jhyap success ($d = 5.022$, $5.321$, $5.567$, all $p < 0.001$). Among non-aggressive agents, ISMCTS showed small but significant advantages over PPO in win rate ($d = 0.343$, $p < 0.001$) and economic performance ($d = 0.682$, $p < 0.001$), and over Random with similar patterns ($d = 0.355$ and $0.639$ respectively, both $p < 0.001$). PPO versus Random showed negligible differences across all metrics (win: $d = 0.017$, $p = 0.704$; economic: $d = -0.076$, $p = 0.088$; Jhyap: $d = 0.043$, $p = 0.333$).

\begin{figure}[H]
    \centering
    \includegraphics[width=\columnwidth]{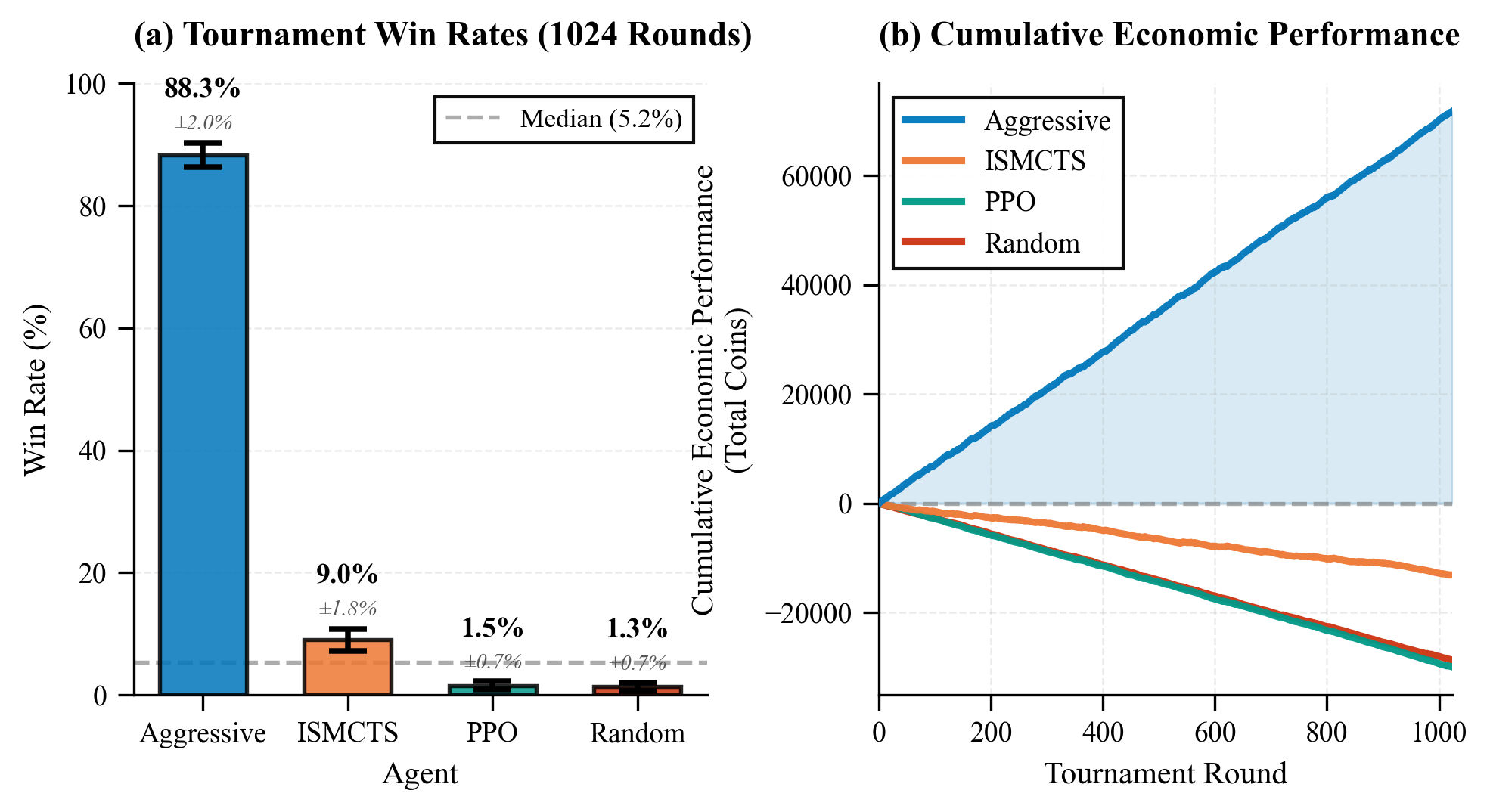}
    \caption{Cross-category championship performance showing (a) tournament win rates with 95\% confidence intervals across all agent categories and (b) cumulative economic performance trajectories over 1024 rounds.}
    \label{fig:championship_plots}
\end{figure}

Figure~\ref{fig:championship_plots} presents the definitive championship evaluation, establishing the Aggressive rule-based agent's categorical superiority across both competitive and economic metrics. Panel (a) reveals a stark performance hierarchy through win rate analysis with 95\% confidence intervals. The Aggressive agent achieves an 88.3\% win rate ($\pm 2.0\%$), representing near-complete dominance over the 1024-round tournament. This performance exceeds the second-place ISMCTS by 79.3 percentage points (ISMCTS: 9.0\% $\pm 1.8\%$), while PPO (1.5\% $\pm 0.7\%$) and Random (1.3\% $\pm 0.7\%$) agents demonstrate statistically indistinguishable performance. The median win rate (5.2\%) illustrates the extreme right-skew of the distribution, with Aggressive operating in a qualitatively distinct performance regime compared to all competitors.

Panel (b) quantifies the economic consequences of these win rate disparities through cumulative coin accumulation trajectories. The Aggressive agent's linear positive trajectory reaches 70.1 coins/round economic performance by round 1024 from Table~\ref{tab:championship_results}. This monotonic accumulation, emphasized by the shaded confidence region, demonstrates systematic profit extraction throughout the tournament without regression phases or vulnerability windows. The trajectory's linearity validates that Aggressive's strategic advantage persists uniformly across game states and opponent configurations, rather than exploiting specific weaknesses that opponents might adapt to or mitigate.

In stark contrast, ISMCTS loses 12.8 per round, while PPO and Random suffer greater losses of 29.2 and 28.0 per round, respectively. The similar decline of PPO and Random indicates that reinforcement learning provides no clear advantage over random play in this multi-agent setting. ISMCTS’s moderate losses suggest that determinized tree search captures some strategic value but remains inferior to Aggressive’s heuristic optimization.

The cumulative results reveal three performance tiers: Aggressive achieves consistent positive gains, ISMCTS incurs moderate losses, and PPO/Random suffer severe economic decline. Aggressive exploits domain-specific strategies, including optimal Jhyap timing, high-value card retention, and aggressive discards. ISMCTS manages imperfect information but with substantial decision overhead, while PPO executes rapidly yet fails to generalize beyond self-play. The large economic gap and exceeding win rate reflect both Aggressive’s frequency of victory and superior coin extraction per win. These findings demonstrate that in Dhumbal, carefully engineered heuristics significantly outperform both search-based and learning-based approaches, emphasizing the critical role of domain-specific strategy in complex multi-agent games.

\newpage
\section{Discussion}\label{sec:discussion}

This section interprets the results of the Dhumbal AI agent evaluation, situates the findings within the context of related work on AI in imperfect-information card games, addresses the limitations of the study, and proposes directions for future research. The unexpected dominance of the rule-based Aggressive agent over sophisticated search-based and learning-based approaches prompts a deeper analysis of strategic effectiveness, computational trade-offs, and the influence of Dhumbal's unique game mechanics on agent performance.

\subsection{Interpretation of Results}

The Aggressive agent's commanding performance in the cross-category championship, achieving an 88.3\% win rate (95\% CI: [86.3, 90.3]), 70.1 coins per round, and 92.6\% Jhyap success rate, underscores the efficacy of simple, risk-tolerant heuristics in Dhumbal. This agent's strategy, which involves frequent Jhyap declarations at hand values $\leq 10$ and prioritization of high-value or multi-card discards, effectively exploits the game's mechanics. In Dhumbal, the ability to end rounds early via Jhyap calls provides a significant advantage in multiplayer settings, where opponents are likely to hold higher-value hands. The Aggressive agent's high economic performance reflects its success in balancing risk and reward, maintaining low hand values while capitalizing on opponents' penalties.

In contrast, the ISMCTS agent, with a 9.0\% win rate (95\% CI: [7.2, 10.8]) and a perfect Jhyap success rate (100\%), demonstrated the strength of belief state modeling in handling imperfect information. By sampling multiple determinizations to estimate opponent hands, ISMCTS made highly accurate Jhyap decisions. However, its negative economic performance (-12.8 coins/round) and substantial decision time (1450.7 ms) highlight computational inefficiencies, likely due to the large action space (128 discretized actions) and the need for multiple determinizations per decision. This contrasts sharply with the Aggressive agent's negligible decision time (0.01 ms), which enables rapid play without sacrificing strategic depth.

The PPO agent's poor performance, with a 1.5\% win rate (95\% CI: [0.8, 2.2]) and -29.2 coins per round, is particularly striking given its superiority over the DQN agent in the learning-based tournament (55.4\% vs. 44.6\% win rate). The Random baseline's dismal performance (1.3\% win rate, -28.0 coins/round) confirms the necessity of strategic decision-making in Dhumbal, as random actions fail to navigate the game's imperfect-information dynamics or optimize economic outcomes.

These results suggest that Dhumbal's mechanics, particularly the Jhyap threshold and multi-card discard options, favor strategies that prioritize aggressive play over complex modeling. The Aggressive agent's success highlights the importance of game-specific heuristics in environments with relatively low information asymmetry (e.g., observable discard pile and hand sizes) compared to games like poker.

\subsection{Comparison with Related Work}

The dominance of a heuristic strategy in Dhumbal contrasts with findings in other imperfect-information card games, where search-based and learning-based methods typically excel. For instance, Brown and Sandholm's Pluribus achieved superhuman performance in multiplayer no-limit Texas Hold'em using a hybrid of Monte Carlo Counterfactual Regret Minimization (MCCFR) and reinforcement learning, leveraging extensive computational resources to model opponent strategies \cite{brown2019}. Similarly, Ginsberg's GIB program for bridge outperformed rule-based approaches by simulating possible card distributions \cite{ginsberg1999}. In Dhumbal, however, the simpler action space (Jhyap, discard, pick) and lower information asymmetry may reduce the advantage of such methods, allowing the Aggressive agent's heuristic to dominate.

The ISMCTS agent's performance aligns with prior work on imperfect-information games. Cowling et al. demonstrated that determinization-based MCTS variants improved decision-making in Magic: The Gathering, a game with complex hidden information \cite{cowling2012mtg}. In Dhumbal, ISMCTS's perfect Jhyap success rate reflects similar strengths in belief state modeling, but its computational overhead limits its effectiveness compared to the Aggressive agent's rapid decisions. This trade-off echoes challenges noted in real-time game settings, where computational efficiency is critical \cite{browne2012}.

The PPO agent's underperformance parallels difficulties observed in reinforcement learning for games with sparse or complex rewards, such as Hanabi \cite{bard2020}. In Hanabi, cooperative dynamics and delayed rewards posed challenges for policy convergence, similar to Dhumbal's competitive multiplayer environment and intricate reward structure.

Dhumbal's multi-card discard mechanics (sets and sequences) introduce additional complexity, making the Aggressive agent's ability to balance risk and opportunity particularly effective. This finding contributes to the literature by demonstrating that heuristic strategies can outperform advanced methods in games with specific structural properties.

\subsection{Limitations}

The study has several limitations that warrant consideration:

\begin{enumerate}
    \item \textbf{Computational Constraints for Search-Based Agents}: ISMCTS's high decision time (1450.7 ms) restricts its applicability in real-time scenarios. The use of only three determinizations per iteration may also limit its robustness in modeling opponent hands, particularly in 4-player games with larger belief spaces.
    \item \textbf{Game-Specific Generalizability}: The Aggressive agent's dominance may be specific to Dhumbal's mechanics, such as the Jhyap threshold ($\leq 10$) and multiplayer dynamics. In games with greater information asymmetry or different win conditions (e.g., poker, bridge), heuristic strategies may be less effective.
    \item \textbf{Simulation Scale}: While 1024 rounds provided sufficient statistical power ($\beta \geq 0.80$ for $\delta \geq 5\%$), a larger sample size (e.g., 50,000 rounds) could improve the precision of win rate estimates, particularly for low-performing agents like PPO and Random.
\end{enumerate}

\subsection{Future Directions}

To address these limitations and extend the study's contributions, the following research directions are proposed:

\begin{enumerate}
    \item \textbf{Generalizability Across Games}: Test the Aggressive agent's performance in variant Dhumbal rules (e.g., Jhyap threshold $\leq 8$, 5–6 players) and other card games (e.g., Yaniv, Rummy) to assess the robustness of heuristic strategies in diverse settings.
    \item \textbf{Human-AI Interaction Studies}: Evaluate the Aggressive agent against human players to identify exploitable weaknesses and develop adaptive heuristics that respond to human strategies, enhancing practical applicability.
    \item \textbf{Multi-Agent Reinforcement Learning}: Explore frameworks like QTRAN \cite{liang2019} to model cooperative and competitive dynamics in Dhumbal, potentially improving learning-based agents' performance in multiplayer settings.
\end{enumerate}

\section{Conclusion}\label{sec:conclusion}

This study evaluates AI agents for Dhumbal, a multiplayer imperfect-information card game, revealing the rule-based Aggressive agent’s dominance with an 88.3\% win rate (95\% CI: [86.3, 90.3]), 70.1 coins/round, and 92.6\% Jhyap success, outperforming the computationally intensive ISMCTS agent (9.0\% win rate, -12.8 coins/round) and the PPO agent (1.5\% win rate, -29.2 coins/round). Contributions include formalizing Dhumbal’s rules, implementing diverse AI agents, conducting statistically rigorous tournaments, and providing open-source code (\url{https://github.com/sahajrajmalla/dhumbal-ai}), highlighting the efficacy of simple heuristics in games with moderate information asymmetry. These findings advance game AI by demonstrating the value of cultural games as testbeds, with future work needed to enhance learning-based agents, improve search efficiency, and test generalizability for broader AI and cultural preservation applications.


\section*{Acknowledgments}
I express my sincere gratitude to my family, and especially my mother, for their steadfast support and encouragement throughout this project.


\bibliographystyle{IEEEtran}   
\bibliography{references}      

\newpage




\vfill

\end{document}